\documentclass[runningheads]{llncs}
\usepackage[T1]{fontenc}
\usepackage{graphicx,verbatim}
\usepackage[utf8]{inputenc}
\usepackage{pifont}
\usepackage{tikz}
\usetikzlibrary{shapes, arrows.meta, positioning, calc}
\usepackage{amsmath, amssymb, amsfonts}    
\usepackage{mathtools}
\usepackage{euscript}
\usetikzlibrary{shadows}
\usepackage{graphicx}
\graphicspath{{Figures/}} 
\usepackage{subcaption}
\usepackage{placeins}
\usepackage{caption} 
\usepackage{float}
\usepackage[table]{xcolor} 
\usepackage{booktabs}
\usepackage{multirow}
\usepackage{adjustbox}
\usepackage{tabularx, array}
\usepackage{colortbl}
\usepackage{float}
\usepackage{algorithm}
\usepackage{algpseudocode} 
\usepackage[most]{tcolorbox} 
\usepackage{verbatim}
\usepackage{comment}
\usepackage[hidelinks]{hyperref}
\usepackage[capitalise,noabbrev]{cleveref}
\hypersetup{
    colorlinks=true,
    linkcolor=blue,
    filecolor=cyan,      
    urlcolor=magenta,
}

\algrenewcommand\algorithmicrequire{\textbf{Input:}}
\algrenewcommand\algorithmicensure{\textbf{Output:}}

\newtcolorbox{algoFrame}{
  colback=white,
  colframe=black!25,
  arc=3pt,
  boxrule=0.6pt,
  left=6pt,right=6pt,top=4pt,bottom=4pt
}
\usepackage{color}

\begin{document}
\title{Echo-CoPilot: A Multiple-Perspective Agentic Framework for Reliable Echocardiography Interpretation}
\titlerunning{Echo-CoPilot}
%

\author{
Moein Heidari\inst{1}\textsuperscript{\textdagger} \and
Ali Mehrabian\inst{1} \and
Mohammad Amin Roohi\inst{1} \and
River Jiang\inst{1} \and
Wenjin Chen\inst{2} \and
David J. Foran\inst{2} \and
Jasmine Grewal\inst{1} \and
Ilker Hacihaliloglu\inst{1}
}

\index{Heidari, Moein}
\index{Mehrabian, Ali}
\index{Roohi, Mohammad Amin}
\index{River Jiang}
\index{Chen, Wenjin}
\index{Foran, David J.}
\index{Grewal, Jasmine}
\index{Hacihaliloglu, Ilker}

\authorrunning{Heidari et al.}

\institute{
The University of British Columbia, Vancouver, BC, Canada \and
Rutgers Cancer Institute, New Brunswick, NJ, USA \\
\email{moein.heidari@ubc.ca}
}
\maketitle              
\begingroup
\renewcommand{\thefootnote}{\fnsymbol{footnote}}
\renewcommand{\thefootnote}{\textdagger}
\footnotetext[1]{Corresponding author}
\endgroup
\begin{abstract}

Echocardiography interpretation requires integrating multi-view temporal evidence with quantitative measurements and guideline-grounded reasoning, yet existing foundation-model pipelines largely solve isolated subtasks and fail when tool outputs are noisy, or values fall near clinical cutoffs. We propose Echo-CoPilot, an end-to-end agentic framework that combines a multi-perspective workflow with knowledge-graph–guided measurement selection. Echo-CoPilot runs three independent ReAct-style agents, structural, pathological, and quantitative, that invoke specialized echocardiography tools to extract parameters while querying EchoKG to determine which measurements are required for the clinical question and which should be avoided. A self-contrast language model then compares the evidence-grounded perspectives, generates a discrepancy checklist, and re-queries EchoKG to apply the appropriate guideline thresholds and resolve conflicts, reducing hallucinated measurement selection and borderline flip-flops. On MIMICEchoQA, Echo-CoPilot provides higher accuracy compared to SOTA baselines and, under a stochasticity stress test, achieves higher reliability through more consistent conclusions and fewer answer changes across repeated runs.  Our code is publicly available at \href{https://github.com/moeinheidari7829/Echo-CoPilot}{GitHub}.



\end{abstract}
\section{Introduction}
Echocardiography interpretation demands clinically reliable decision-making by integrating multi-view temporal evidence, quantitative measurements, and strict alignment with clinical guidelines \cite{mitchell2019guidelines}. While deep learning has achieved strong performance on isolated perceptual tasks such as view classification, segmentation, and functional quantification, full-study interpretation remains a bottleneck.
Large language models (LLMs) and agentic systems offer a promising paradigm by treating interpretation as a procedure, dynamically invoking specialized perceptual tools to synthesize a final answer \cite{yao2022react}. 
However, reliance on a single reasoning trajectory may reduce robustness in clinical settings, particularly near severity thresholds, where small variations in measurements can alter the diagnostic conclusion \cite{hager2024evaluation}. This sensitivity is further amplified in echocardiography, where conclusions depend on specific views and guideline criteria, and where tool outputs may be noisy or occasionally inconsistent across views \cite{unlu2020impact}.
To address this, we propose Echo-CoPilot, a multi-perspective agentic framework for echocardiography interpretation that improves both performance and reliability by explicitly contrasting complementary clinical reasoning pathways. Our main contributions are: 
\ding{202}\ \textbf{\textit{EchoKG}}: We introduce the Echocardiography knowledge-graph (EchoKG), a clinical knowledge-graph derived from consensus echocardiography guidelines, capturing diagnostic dependencies and formalizing measurement selection and avoidance rules as structured inference-time constraints.
\ding{203}\ \textbf{\textit{Multiple-Perspective Framework}}: We propose a multi-perspective procedure. Echo-CoPilot instantiates three parallel ReAct agents, Structural, Pathological, and Quantitative, that independently reason over the exam using a shared tool cache that reuses tool output to improve efficiency. A self-contrast LLM then aggregates these diverse trajectories, generating a discrepancy checklist to resolve conflicts while ensuring the final decision remains constrained by EchoKG.
\ding{204}\ \textbf{\textit{Performance Evaluation}}: We achieve state-of-the-art (SOTA) accuracy on the MIMICEchoQA benchmark and empirically demonstrate that our agentic framework improves diagnostic stability and reduces hallucination and reasoning variance compared to single-trajectory baselines.

\section{Related Work}
\textbf{Vision Foundation Models in Echocardiography.} 
While early methods focused on narrow, task-specific supervision \cite{ouyang2020video}, recent efforts leverage scalable foundation models. Vision models like EchoApex \cite{amadou2024echoapex} and MedSAM2 \cite{ma2025medsam2} provide robust representation and promptable segmentation, while vision-language models (VLMs) like PanEcho \cite{holste2025complete} and EchoPrime \cite{vukadinovic2024echoprime} map entire video studies to clinical descriptors or reports. However, these end-to-end models operate as "black boxes" \cite{rajpurkar2022ai}, lacking the multi-step reasoning needed to decompose complex clinical queries or audit their logic against medical guidelines.

\textbf{Agentic Frameworks in Medical Imaging.} 
Tool augmented LLM agents aim to increase transparency by decomposing a task and invoking specialist tools \cite{yao2022react}. Radiology agents like MedRAX \cite{fallahpour2025medraxmedicalreasoningagent} decompose X-ray interpretation into verifiable steps, but transferring this paradigm to echocardiography requires handling complex temporal dynamics and view-dependent feasibility. While concurrent works like EchoAgent \cite{daghyani2025echoagent} explore tool-assisted measurements, their single-trajectory reasoning cannot resolve the cross-tool and cross-view disagreements that are intrinsic to temporal echocardiographic evidence.

\textbf{Constrained and verifiable reasoning.}
Recent work improves reliability through self-critique and external grounding, including reflective loops and structured knowledge integration \cite{huang2025medreflect,qin2025multi}. Echo-CoPilot builds on these directions but targets echocardiography-specific failure modes. We use our proposed EchoKG to enforce measurement selection guided by the question intent and avoidance at inference time, and propose a multiple-perspective procedure \cite{zhang2024selfcontrast} to surface inconsistencies and synthesize a checklist-guided final answer grounded in tool evidence.
\section{Method}
\label{sec:method}
\begin{figure}[t]  
  \centering
  \includegraphics[width=0.81\textwidth]{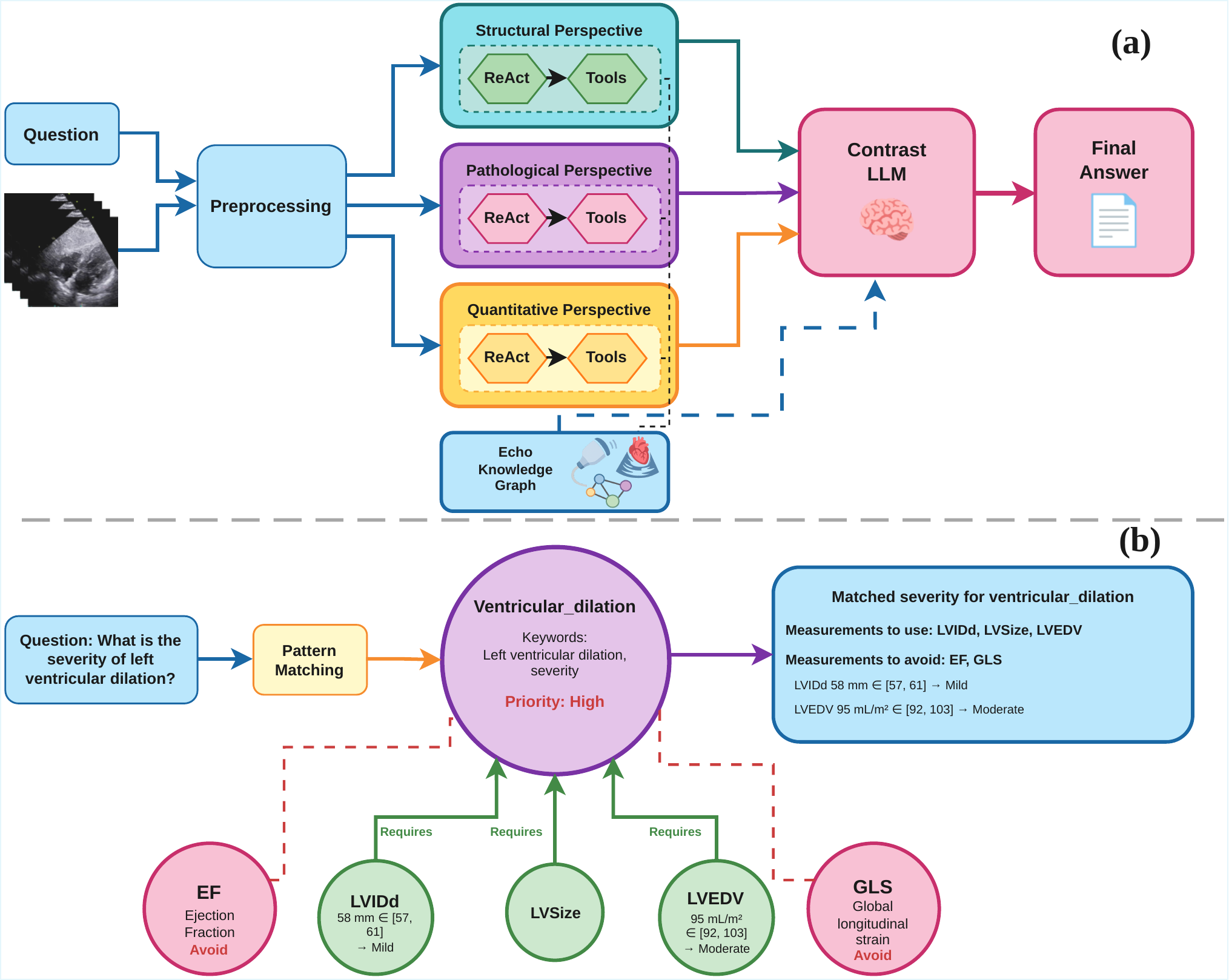}  
  \caption{Overview of Echo-CoPilot. Panel (a) shows the three perspective ReAct-style agents and the contrast module. Panel (b) illustrates EchoKG measurement selection/avoidance and threshold-based grading of ventricular dilation severity.}
  \label{fig:method}  
\end{figure}

Echo-CoPilot, an LLM-driven agentic system, decomposes
a query into actionable steps, selectively invokes specialized tools, integrates intermediate findings, and synthesizes a transparent, clinically aligned final
assessment. As illustrated in Figure~\ref{fig:method}, our proposed framework employs a two-stage architecture for robust echo interpretation: (1) multiple-perspective generation and (2) self-contrast analysis. Our proposed EchoKG is shared between these two stages to leverage an additional source of knowledge and reduce hallucinations.
\\
\textbf{\textit{Multi-Perspective Generation.}}
Inspired by \cite{zhang2024selfcontrast}, given a question $q$ and video $v$, we generate three independent perspectives using specialized system prompts:

\begin{itemize}
    \item \textbf{Structural Perspective} ($P_1$): Focuses on anatomical structures, morphology, and chamber dimensions.
    \item \textbf{Pathological Perspective} ($P_2$): Emphasizes disease patterns and clinical indicators.
    \item \textbf{Quantitative Perspective} ($P_3$): Prioritizes numerical measurements and quantitative thresholds.
\end{itemize}

Each perspective executes an independent ReAct loop \cite{yao2022react}, autonomously selecting its specific tools (e.g., echo measurement prediction, echo disease prediction) and producing a final answer. The EchoKG is also shared between all agents to use the clinical knowledge and thresholds. We design a shared deterministic tool cache in which identical tool calls return the same cached output, eliminating redundancy while preserving consistency. The multi-perspective approach reduces hallucination by generating independent analyses that can cross-validate each other, catching errors that a single perspective might miss.
\\
Echo-CoPilot is equipped with a set of domain-specific echocardiography tools, each implemented
as a callable module with a structured input schema and standardized output format. 
In our implementation, the core tool set includes the EchoPrime model for view classification and measurement prediction \cite{vukadinovic2024echoprime}, PanEcho for disease and finding prediction \cite{holste2025complete}, report style synthesis \cite{vukadinovic2024echoprime}, and our proposed EchoKG. Crucially, the reasoning agents are LLM-driven, while perception is delegated to dedicated vision–language tools.

\textbf{\textit{Self-Contrast Mechanism.}}
The contrast phase employs a dedicated contrast agent that analyzes the three perspectives $P=\{P_1, P_2, P_3\}$ and generates a structured discrepancy checklist. The contrast prompt includes: (1) all perspective responses, (2) aggregated measurements from tool outputs, (3) EchoKG guidance and clinical thresholds retrieved via $\text{EchoKG}(q)$, and (4) validation instructions requiring the model to produce an explicit discrepancy checklist before its final answer.
The contrast LLM first generates a structured discrepancy checklist by validating whether perspectives followed EchoKG guidance, identifying severity classification discrepancies, comparing quantitative measurement consistency, and assessing tool reliability. The contrast LLM then synthesizes a final answer that addresses all checklist items, resolves discrepancies, and uses EchoKG thresholds to determine severity classifications. The final answer integrates evidence from all perspectives while giving higher weight to perspectives that followed EchoKG guidance. The contrast phase explicitly validates measurement selection against EchoKG guidance, giving higher weight to perspectives that followed medical principles rather than simple majority voting.

\textbf{\textit{Echocardiography knowledge-graph.}}
To enforce clinically sound reasoning and prevent metric hallucination, we propose the EchoKG. As illustrated in \cref{fig:method} (Panel (b)), EchoKG is a directed graph $G = (V, E)$ that uses guideline logic as queryable constraints via three node types: (1) \textbf{Structure nodes} ($V_s$) representing anatomical components, (2) \textbf{Measurement nodes} ($V_m$) defining standard metrics (e.g., LVEDV, EF), and (3) \textbf{Pattern nodes} ($V_p$) comprising 16 distinct clinical query templates (e.g., cavity dilation, valvular regurgitation). EchoKG governs deterministic measurement selection by mapping Pattern nodes to Measurement nodes via strict \textit{requires} and \textit{avoid} directed edges. The graph topology and mapping rules are meticulously distilled from major clinical guidelines (e.g., ASE, EACVI \cite{mitchell2019guidelines,valsangiacomo2015indications,unlu2020impact}\footnote{Incorporated materials are utilized strictly for knowledge construction and inference-time retrieval rather than model training, in full compliance with their terms of use.}), serving as a guardrail for the agent.
\\
Given question $q$, $\text{EchoKG}(q)$ executes: (1) \textbf{Pattern matching}: Patterns are evaluated by priority (1=most specific, 3=most general), ensuring specific patterns (e.g., atrial enlargement) match before general ones (e.g., cavity dilation), reducing false positives. For pattern $p$ with keywords $K$, context $K_c$, and exclusions $K_e$, $p$ matches if $K \cap q \neq \emptyset$, $K_c \cap q \neq \emptyset$ (if $K_c$ exists), and $K_e \cap q = \emptyset$. (2) \textbf{Structure disambiguation}: If $p$ supports multiple structures (stored as metadata), detect the specific structure from $q$ via keyword matching. (3) \textbf{Graph traversal}: Follow "requires" edges from $p$ to get $M_{\text{req}}$ and "avoid" edges to get $M_{\text{avoid}}$. Unlike traditional knowledge graphs with only positive edges, EchoKG employs dual-edge relationships with explicit "avoid" edges providing negative guidance (e.g., avoid EF for cavity size questions), preventing common measurement selection errors. (4) \textbf{Threshold retrieval}: Extract clinical thresholds (e.g., normal ranges, severity classifications) from measurement node attributes. EchoKG stores clinical thresholds in measurement nodes to provide guideline-based reference values and reduce hallucinated ranges. It is queried during perspective generation and contrast analysis with a shared cache for consistent guidance. The proposed framework is summarized in \cref{alg:multiperspective-contrast}.
\begin{algorithm}[t]
\caption{Proposed Echo-CoPilot Agentic Framework}
\label{alg:multiperspective-contrast}
\footnotesize
\begin{algorithmic}[1]
\Require Question $q$, video $v$, tools $T$, perspective prompts $\{\text{prompt}_i\}_{i=1}^{3}$
\Ensure Final answer $A_{\text{final}}$
\State $C \gets \emptyset$ \Comment{Shared cache for deterministic tool calls}
\State $g \gets \textsc{EchoKG}(q)$ \Comment{Measurement selection and avoidance rules available to all stages}
\For{$i \in \{1,2,3\}$}
  \State $P_i \gets \textsc{ReAct}(q,v,\text{prompt}_i,T;\,C, g)$
  \Comment{$P_i$ includes answer and tool evidence, guided by $g$}
\EndFor
\State $m \gets \textsc{Aggregate}(\{P_i\}_{i=1}^{3})$ \Comment{Measurements from tool outputs}
\State $\mathcal{L} \gets \textsc{ContrastLLM}(\{P_i\}, m, g)$ \Comment{Discrepancy checklist with EchoKG constraints}
\State $A_{\text{final}} \gets \textsc{Refine}(\{P_i\}, \mathcal{L}, m, g)$ \Comment{Final constrained resolution}
\State \Return $A_{\text{final}}$
\end{algorithmic}
\end{algorithm}
\section{Experiments}
\label{sec:experiments}
\textbf{\textit{Dataset, Implementation Details, and Evaluation Metrics.}}
We evaluate Echo-CoPilot on MIMICEchoQA \cite{thapa2025well}, which contains 622 transthoracic echocardiogram videos paired with multiple-choice questions (four options: A–D) spanning 38 standard views. The questions cover core interpretation tasks, including ventricular systolic function, chamber size, valvular stenosis, and pericardial effusion. As this benchmark is strictly reserved for zero-shot evaluation, our agentic framework had no exposure to it during training; we report accuracy on the entire held-out dataset \cite{thapa2025well}. Echo-CoPilot uses open-source gpt-oss-120b \cite{agarwal2025gpt} as the LLM and invokes tools through structured function calls implemented with the LangChain/LangGraph framework. Videos are preprocessed by frame sampling and resolution normalization to ensure stable cross-tool compatibility. All experiments are run on a single NVIDIA A100 (80GB) GPU. Other implementation choices include the 3-perspective prompts and LLM temperatures (contrast=0.3, perspective=0.7). For image-only baselines such as MedGemma 4B \cite{sellergren2025medgemma}, we follow prior practice and evaluate on key frames extracted from each clip, since these models are not designed to ingest full video.
For the specific analysis of system stability and agentic reasoning consistency, we created a stratified subset of 50 challenging question-answer pairs covering diverse categories (structural, functional, and disease-related queries). The maximum reasoning steps for the ReAct loop were set to 10.
\begin{table}[t]
\centering
\caption{Accuracy (Acc.) of general-purpose and biomed-specialized VLMs on the MIMICEchoQA benchmark. Models marked with \textsuperscript{*} are taken directly from OpenBiomedVid~\cite{thapa2025well,thapa2026how}.}
\label{tab:mimic_echoqa}
\setlength{\tabcolsep}{6pt}
\renewcommand{\arraystretch}{1.12}
\small
\begin{tabularx}{\textwidth}{@{}X c X c@{}}
\toprule
\textbf{Model} & \textbf{Acc.} & \textbf{Model} & \textbf{Acc.} \\
\midrule
Video-ChatGPT\textsuperscript{*} & 31.7 & Video-LLaVA\textsuperscript{*} & 32.0 \\
Phi-3.5-Vision-Instruct\textsuperscript{*} & 41.1 & Phi-4-Multimodal-Instruct\textsuperscript{*} & 37.8 \\
InternVideo2.5-Chat-8B\textsuperscript{*} & 40.3 & Qwen2-VL-7B-Instruct\textsuperscript{*} & 37.9 \\
Qwen2.5-VL-7B-Instruct\textsuperscript{*} & 34.0 & Qwen2 VL 72B Instruct\textsuperscript{*} & 37.5 \\
Qwen2.5-VL-72B-Instruct\textsuperscript{*} & 34.2 & Gemini-2.0-Flash\textsuperscript{*} & 38.4 \\
GPT-4o\textsuperscript{*} & 41.6 & GPT-o4-mini\textsuperscript{*} & 43.9 \\
Qwen2-VL-2B-Biomed\textsuperscript{*} & 42.0 & Qwen2-VL-7B-Biomed\textsuperscript{*} & 49.0 \\
MedGemma-4B & 33.4 & LLaVA-Med & 32.5 \\
\midrule
\rowcolor{gray!12}
\textbf{Echo-CoPilot} & \textbf{54.0\,$\pm$\,0.7} & & \\
\bottomrule
\end{tabularx}
\end{table}
To validate the efficacy and robustness of Echo-CoPilot, our experiments focus on two key dimensions: (1) diagnostic accuracy compared to SOTA VLMs, and (2) response stability and consistency, demonstrating the specific value of the proposed multiple-perspective mechanism.
\subsection{Diagnostic Accuracy}
Table~\ref{tab:mimic_echoqa} reports accuracy on MIMICEchoQA. Echo-CoPilot achieves the best overall performance, outperforming both proprietary and open-source multimodal baselines. The gains are largest on questions that require quantitative reasoning or integrating multiple cues, such as grading systolic dysfunction from ejection fraction, assessing hypertrophy severity, or stratifying pericardial effusion. In these cases, vision-only models often rely on appearance-driven heuristics and are more likely to flip labels near guideline cutoffs, whereas Echo-CoPilot defers to exam-specific measurements and disease cues produced by its tools and then applies guideline-grounded interpretation through EchoKG. 
\Cref{fig:sc_example} shows a representative failure case where the value of the self-contrast (SC) mechanism is clear. For the query “Is there aortic stenosis?” (ground truth: No), one perspective incorrectly treated the tool confidence (84\%) as the disease probability and voted Yes, while the other two perspectives correctly voted No based on the categorical tool output (“No stenosis detected”). The SC layer detected this conflict, used EchoKG to prioritize the categorical label, and recovered the correct diagnosis.

\begin{figure}[!t]
\centering
\resizebox{0.9\textwidth}{!}{%
\begin{tikzpicture}[
    font=\small\sffamily,
    box/.style={draw, rounded corners, align=center, inner sep=6pt, drop shadow},
    input/.style={box, fill=gray!5, rectangle, minimum width=2.8cm, minimum height=3.2cm},
    correct/.style={box, fill=green!5, draw=green!60!black, thick, minimum width=2.8cm, minimum height=1.8cm},
    error/.style={box, fill=red!5, draw=red!60!black, thick, minimum width=2.8cm, minimum height=1.8cm},
    agg/.style={box, fill=blue!5, draw=blue!60!black, thick, minimum width=3.4cm, minimum height=3.6cm},
    arrow/.style={-Latex, thick, color=gray!70!black}
]

\node (input) [input] {
    \textbf{Query:}\\
    ``Is there aortic\\
    stenosis?''\\[3pt]
    \textbf{Ground Truth:} No\\[3pt]
    \textit{Tool Output:}\\
    \texttt{\footnotesize NO STENOSIS}\\
    \texttt{\footnotesize DETECTED}\\
    \texttt{\footnotesize (Conf: 84\%)}
};

\node (struct) [correct, right=2.5cm of input, yshift=2.8cm] {
    \textbf{P1: Structural}\\
    \rule{2.4cm}{0.4pt}\\
    {\scriptsize\textit{Reasoning:}}\\
    {\scriptsize Relies on categorical}\\
    {\scriptsize ``No Stenosis'' label.}\\[2pt]
    \textbf{Vote: No} \textcolor{green!60!black}{\ding{51}}
};

\node (patho) [error, right=2.5cm of input, yshift=0cm] {
    \textbf{P2: Pathological}\\
    \rule{2.4cm}{0.4pt}\\
    {\scriptsize\textit{Reasoning Error:}}\\
    {\scriptsize Misinterprets 84\%}\\
    {\scriptsize conf.\ as prob.\ of \textit{presence}.}\\[2pt]
    \textbf{Vote: Yes} \textcolor{red}{\ding{55}}
};

\node (quant) [correct, right=2.5cm of input, yshift=-2.8cm] {
    \textbf{P3: Quantitative}\\
    \rule{2.4cm}{0.4pt}\\
    {\scriptsize\textit{Reasoning:}}\\
    {\scriptsize Aligns tool output}\\
    {\scriptsize with EchoKG thresholds.}\\[2pt]
    \textbf{Vote: No} \textcolor{green!60!black}{\ding{51}}
};

\node (agg) [agg, right=2.5cm of patho] {
    \textbf{Self-Contrast}\\
    \textbf{Mechanism}\\[3pt]
    {\scriptsize\textit{Obs:} P2 contradicts}\\
    {\scriptsize P1, P3, \& EchoKG.}\\[2pt]
    {\scriptsize\textit{Res:} Corrects}\\
    {\scriptsize the hallucination.}\\[3pt]
    \textbf{Consensus: No}
};

\node (output) [correct, right=2cm of agg, minimum width=2.4cm, minimum height=1.4cm, fill=green!10] {
    \textbf{Final Decision:}\\
    \textbf{No} \textcolor{green!60!black}{\ding{51}}
};

\draw[arrow] (input.east) -- ++(0.7,0) |- (struct.west);
\draw[arrow] (input.east) -- (patho.west);
\draw[arrow] (input.east) -- ++(0.7,0) |- (quant.west);

\draw[arrow] (struct.east) -| ([xshift=-1.25cm, yshift=0.4cm]agg.west) -- ([yshift=0.4cm]agg.west);
\draw[arrow] (patho.east) -- (agg.west);
\draw[arrow] (quant.east) -| ([xshift=-1.25cm, yshift=-0.4cm]agg.west) -- ([yshift=-0.4cm]agg.west);

\draw[arrow] (agg.east) -- (output.west);

\end{tikzpicture}
}
\caption{Multiple-Perspective Error Correction. When the deterministic tool output is misinterpreted by a single reasoning perspective (e.g., P2), the self-contrast mechanism successfully isolates and corrects the logic error.}
\label{fig:sc_example}
\end{figure}
\begin{table}[!ht]
\centering
\caption{Stability Comparison across Testing Modes: consistency over 10 runs for $N{=}50$ questions. Best results are in \textbf{bold.}}
\label{tab:stability}
\setlength{\tabcolsep}{7pt}
\begin{adjustbox}{width=\textwidth}
\begin{tabular}{@{}lcccc@{}}
\toprule
\textbf{Metric} & \textbf{LLM} & \textbf{LLM+Tools} & \textbf{ReAct Agent} & \textbf{Echo-CoPilot} \\
\midrule
Stability Rate $\uparrow$ & 40.0\% & 26.0\% & 70.0\% & \textbf{72.0\%} \\
Avg Unique Answers $\downarrow$ & 1.86 & 2.00 & 1.30 & \textbf{1.22} \\
Avg Changes $\downarrow$ & 2.10 & 2.58 & 1.14 & \textbf{0.96} \\
\bottomrule
\end{tabular}
\end{adjustbox}
\end{table}
\subsection{Stability and Consistency Analysis, Ablation Study.}
\noindent
Clinical deployment demands reproducible decisions under stochastic decoding and noisy tool outputs \cite{shyr2025statistical}. We evaluate stability across 10 runs on 50 questions ($500$ total inferences) under four configurations: LLM, LLM+Tools (naive function calling), ReAct Agent, and Echo-CoPilot. Following \cite{bouchard2025uncertainty}, we measure \emph{Stability Rate} (fraction of identical predictions across 10 runs), \emph{Avg Unique Answers}, and \emph{Avg Changes} (mean number of answer flips). \Cref{tab:stability} shows that naive tool access severely degrades stability (26\%) by amplifying uncontextualized tool noise. Conversely, the ReAct agent recovers stability (70\%) by structuring evidence accumulation, while our proposed Echo-CoPilot with the SC mechanism further suppresses fluctuations (0.96 Avg Changes), achieving the highest overall stability (72\%). This confirms that the SC mechanism effectively resolves borderline cases by grounding perspective-specific evidence in shared EchoKG logic.

\begin{table}[!t]
\centering
\caption{Left: Error concentration on three frequent failure groups, comparing Qwen2-VL-7B-Biomed \cite{thapa2026how} with Echo-CoPilot. Right: Ablation study on each component of Echo-CoPilot: EchoKG, ReAct loop, and self-contrast (SC) mechanism. Best results are in \textbf{bold}.}
\label{tab:error_ablation}
\footnotesize
\setlength{\tabcolsep}{2.6pt}
\begin{tabular}{@{}lcc@{\hskip 6pt}||@{\hskip 6pt}lccccr@{}}
\toprule
\multicolumn{3}{c}{\textbf{Error Analysis (err.\%)}} &
\multicolumn{6}{c}{\textbf{Ablation Study}} \\
\cmidrule(lr){1-3}\cmidrule(lr){4-9}
\textbf{Group} & \textbf{Qwen} & \textbf{Ours} &
\textbf{Variant} & \textbf{Tools} & \textbf{EchoKG} & \textbf{ReAct} & \textbf{SC} & \textbf{Acc.} \\
\midrule
Doppler views & 83.0 & \textbf{59.3} & LLM         & $\circ$   & $\circ$   & $\circ$   & $\circ$   & 46.9 \\
PSAX vessels  & 71.0 & \textbf{59.6} & LLM+Tools    & $\bullet$ & $\bullet$ & $\circ$   & $\circ$   & 49.7 \\
Left atrium   & 68.4 & \textbf{60.5} & ReAct Agent  & $\bullet$ & $\bullet$ & $\bullet$ & $\circ$   & 52.3 \\
              &      &               & Echo-CoPilot & $\bullet$ & $\bullet$ & $\bullet$ & $\bullet$ & \textbf{54.0} \\
\bottomrule
\end{tabular}
\end{table}
\noindent
Next, we conduct an error analysis by anatomy and views that exhibit the highest errors \cite{thapa2026how}. \Cref{tab:error_ablation} (Left) demonstrates that our framework significantly mitigates the catastrophic failures of baselines in challenging regions. While the Qwen2-VL-7B-Biomed \cite{thapa2026how} suffers error rates of 83\% and 71\% on the Doppler views and PSAX great vessels, respectively, due to acoustic dropout, our tool-routed agent reduces these to 59.3\% and 59.6\%, proving highly robust against ambiguous acoustic windows.

\noindent Finally, the component ablations (\cref{tab:error_ablation}, Right) show a steady improvement as components are added. The pure LLM baseline reaches 46.9\% accuracy, adding tools and EchoKG (LLM+Tools) increases accuracy to 49.7\%, and incorporating the ReAct loop further to derive an agentic framework (ReAct Agent) improves the performance to 52.3\%. Finally, Echo-CoPilot with the SC mechanism gives the best result, which is a substantial improvement over the pure LLM baseline.

\noindent\textbf{Limitations and the Challenge of Tool Variability.} 
While our proposed Echo-CoPilot framework outperforms state-of-the-art baselines (\cref{tab:mimic_echoqa}), the overall performance ceiling reflects the severe vulnerability of current perceptual tools to domain shifts. We accept this modest accuracy margin as a necessary trade-off for explainable reasoning and verifiable guideline adherence. Notably, our 500-iteration stability test revealed that pure LLM logic errors are remarkably rare ($<1\%$). Instead, \textit{tool variability} remains the primary bottleneck: when foundation models extract noisy or conflicting perceptual data, the self-contrast mechanism exposes cross-perspective disagreements, leading to reasoning that is internally consistent but ultimately factually incorrect. Thus, while our architecture resolves the reasoning bottleneck, future work in agentic echocardiography must prioritize the deterministic reliability of underlying measurement tools.

\section{Conclusion}
We introduced Echo-CoPilot, an agentic framework that advances echocardiography interpretation from black-box prediction to guideline-constrained, verifiable reasoning. By enforcing clinical validity through EchoKG and resolving evidentiary conflicts via multi-perspective and self-contrast mechanisms, Echo-CoPilot achieves SOTA performance on MIMICEchoQA, significantly improving reliability near critical decision boundaries. While the scarcity of public video-based benchmarks currently limits evaluation to a single large-scale cohort, this work establishes a rigorous paradigm for auditable medical agents. Future efforts will focus on curating diverse, multi-center benchmarks to test generalization and advancing toward prospective clinical validation.

\subsubsection{Acknowledgments.}
This work was supported by the Canadian Foundation for Innovation-John R. Evans Leaders Fund (CFI-JELF) program grant number 42816. Mitacs Accelerate program grant number AWD024298-IT33280. We also acknowledge the support of the Natural Sciences and Engineering Research Council of Canada (NSERC), [RGPIN-2023-03575]. Cette recherche a été financée par le Conseil de recherches en sciences naturelles et en génie du Canada (CRSNG), [RGPIN-2023-03575].

\subsubsection{Disclosure of Interests.} The authors have no competing interests in the paper.

\bibliographystyle{splncs04}
\bibliography{Paper-20}

\end{document}